%% file: main.tex
\documentclass[runningheads]{llncs}

\usepackage{eccv}

\usepackage{eccvabbrv}

\usepackage{graphicx}
\usepackage{booktabs}
\usepackage{comment}
\usepackage{multirow}
\usepackage{siunitx}
\usepackage{utfsym}

\usepackage[accsupp]{axessibility}  %

\input{matplotlib_colors}
\input{utils}

\usepackage{hyperref}

\usepackage{orcidlink}

\begin{document}

\title{Massively Multi-Person 3D Human Motion Forecasting with Scene Context}

\author{Felix B Mueller\inst{1}\thanks{work done while at University of Bonn.}\orcidlink{0009-0002-5848-5250} \and
Julian Tanke\inst{2} \and
Juergen Gall\inst{2,3}\orcidlink{0000-0002-9447-3399}}

\authorrunning{Mueller, Tanke, et al.}

\institute{Institute of Computer Science and Campus Institute Data Science, University of Goettingen, Germany \and Institute of Computer Science, University of Bonn, Germany \and Lamarr Institute for Machine Learning
and Artificial Intelligence, Germany\\ \email{felix.mueller@cs.uni-goettingen.de} 
}

\maketitle

\setlength{\tabcolsep}{4pt}

\begin{abstract}
  \input{sections/1_abstract}

\end{abstract}

\section{Introduction}
\label{sec:intro}

\input{sections/2_introduction}

\section{Related Work}
\label{sec:related-work}

\input{sections/2.5_related_work}

\section{Methodology}
\label{sec:methodology}
\input{sections/3_methodology}

\section{Experiments}
\label{sec:experiments}

\input{sections/4_experiments}

\section{Conclusion}
\label{sec:conclusion}

\input{sections/5_conclusion}

\bibliographystyle{splncs04}
\bibliography{main}

\appendix

\section{Example Videos}
\input{suppmat/videos}

\section{Implementation Details}
\input{suppmat/implementation_details}

\section{Additional Metrics}
\input{suppmat/additional_metrics}

\end{document}

%% file: matplotlib_colors.tex
\definecolor{mplaliceblue}{rgb}{0.941176470588,0.972549019608,1.0}
\definecolor{mplantiquewhite}{rgb}{0.980392156863,0.921568627451,0.843137254902}
\definecolor{mplaqua}{rgb}{0.0,1.0,1.0}
\definecolor{mplaquamarine}{rgb}{0.498039215686,1.0,0.83137254902}
\definecolor{mplazure}{rgb}{0.941176470588,1.0,1.0}
\definecolor{mplbeige}{rgb}{0.960784313725,0.960784313725,0.862745098039}
\definecolor{mplbisque}{rgb}{1.0,0.894117647059,0.76862745098}
\definecolor{mplblack}{rgb}{0.0,0.0,0.0}
\definecolor{mplblanchedalmond}{rgb}{1.0,0.921568627451,0.803921568627}
\definecolor{mplblue}{rgb}{0.0,0.0,1.0}
\definecolor{mplblueviolet}{rgb}{0.541176470588,0.16862745098,0.886274509804}
\definecolor{mplbrown}{rgb}{0.647058823529,0.164705882353,0.164705882353}
\definecolor{mplburlywood}{rgb}{0.870588235294,0.721568627451,0.529411764706}
\definecolor{mplcadetblue}{rgb}{0.372549019608,0.619607843137,0.627450980392}
\definecolor{mplchartreuse}{rgb}{0.498039215686,1.0,0.0}
\definecolor{mplchocolate}{rgb}{0.823529411765,0.411764705882,0.117647058824}
\definecolor{mplcoral}{rgb}{1.0,0.498039215686,0.313725490196}
\definecolor{mplcornflowerblue}{rgb}{0.392156862745,0.58431372549,0.929411764706}
\definecolor{mplcornsilk}{rgb}{1.0,0.972549019608,0.862745098039}
\definecolor{mplcrimson}{rgb}{0.862745098039,0.078431372549,0.235294117647}
\definecolor{mplcyan}{rgb}{0.0,1.0,1.0}
\definecolor{mpldarkblue}{rgb}{0.0,0.0,0.545098039216}
\definecolor{mpldarkcyan}{rgb}{0.0,0.545098039216,0.545098039216}
\definecolor{mpldarkgoldenrod}{rgb}{0.721568627451,0.525490196078,0.043137254902}
\definecolor{mpldarkgray}{rgb}{0.662745098039,0.662745098039,0.662745098039}
\definecolor{mpldarkgreen}{rgb}{0.0,0.392156862745,0.0}
\definecolor{mpldarkgrey}{rgb}{0.662745098039,0.662745098039,0.662745098039}
\definecolor{mpldarkkhaki}{rgb}{0.741176470588,0.717647058824,0.419607843137}
\definecolor{mpldarkmagenta}{rgb}{0.545098039216,0.0,0.545098039216}
\definecolor{mpldarkolivegreen}{rgb}{0.333333333333,0.419607843137,0.18431372549}
\definecolor{mpldarkorange}{rgb}{1.0,0.549019607843,0.0}
\definecolor{mpldarkorchid}{rgb}{0.6,0.196078431373,0.8}
\definecolor{mpldarkred}{rgb}{0.545098039216,0.0,0.0}
\definecolor{mpldarksalmon}{rgb}{0.913725490196,0.588235294118,0.478431372549}
\definecolor{mpldarkseagreen}{rgb}{0.560784313725,0.737254901961,0.560784313725}
\definecolor{mpldarkslateblue}{rgb}{0.282352941176,0.239215686275,0.545098039216}
\definecolor{mpldarkslategray}{rgb}{0.18431372549,0.309803921569,0.309803921569}
\definecolor{mpldarkslategrey}{rgb}{0.18431372549,0.309803921569,0.309803921569}
\definecolor{mpldarkturquoise}{rgb}{0.0,0.807843137255,0.819607843137}
\definecolor{mpldarkviolet}{rgb}{0.580392156863,0.0,0.827450980392}
\definecolor{mpldeeppink}{rgb}{1.0,0.078431372549,0.576470588235}
\definecolor{mpldeepskyblue}{rgb}{0.0,0.749019607843,1.0}
\definecolor{mpldimgray}{rgb}{0.411764705882,0.411764705882,0.411764705882}
\definecolor{mpldimgrey}{rgb}{0.411764705882,0.411764705882,0.411764705882}
\definecolor{mpldodgerblue}{rgb}{0.117647058824,0.564705882353,1.0}
\definecolor{mplfirebrick}{rgb}{0.698039215686,0.133333333333,0.133333333333}
\definecolor{mplfloralwhite}{rgb}{1.0,0.980392156863,0.941176470588}
\definecolor{mplforestgreen}{rgb}{0.133333333333,0.545098039216,0.133333333333}
\definecolor{mplfuchsia}{rgb}{1.0,0.0,1.0}
\definecolor{mplgainsboro}{rgb}{0.862745098039,0.862745098039,0.862745098039}
\definecolor{mplghostwhite}{rgb}{0.972549019608,0.972549019608,1.0}
\definecolor{mplgold}{rgb}{1.0,0.843137254902,0.0}
\definecolor{mplgoldenrod}{rgb}{0.854901960784,0.647058823529,0.125490196078}
\definecolor{mplgray}{rgb}{0.501960784314,0.501960784314,0.501960784314}
\definecolor{mplgreen}{rgb}{0.0,0.501960784314,0.0}
\definecolor{mplgreenyellow}{rgb}{0.678431372549,1.0,0.18431372549}
\definecolor{mplgrey}{rgb}{0.501960784314,0.501960784314,0.501960784314}
\definecolor{mplhoneydew}{rgb}{0.941176470588,1.0,0.941176470588}
\definecolor{mplhotpink}{rgb}{1.0,0.411764705882,0.705882352941}
\definecolor{mplindianred}{rgb}{0.803921568627,0.360784313725,0.360784313725}
\definecolor{mplindigo}{rgb}{0.294117647059,0.0,0.509803921569}
\definecolor{mplivory}{rgb}{1.0,1.0,0.941176470588}
\definecolor{mplkhaki}{rgb}{0.941176470588,0.901960784314,0.549019607843}
\definecolor{mpllavender}{rgb}{0.901960784314,0.901960784314,0.980392156863}
\definecolor{mpllavenderblush}{rgb}{1.0,0.941176470588,0.960784313725}
\definecolor{mpllawngreen}{rgb}{0.486274509804,0.988235294118,0.0}
\definecolor{mpllemonchiffon}{rgb}{1.0,0.980392156863,0.803921568627}
\definecolor{mpllightblue}{rgb}{0.678431372549,0.847058823529,0.901960784314}
\definecolor{mpllightcoral}{rgb}{0.941176470588,0.501960784314,0.501960784314}
\definecolor{mpllightcyan}{rgb}{0.878431372549,1.0,1.0}
\definecolor{mpllightgoldenrodyellow}{rgb}{0.980392156863,0.980392156863,0.823529411765}
\definecolor{mpllightgreen}{rgb}{0.564705882353,0.933333333333,0.564705882353}
\definecolor{mpllightgrey}{rgb}{0.827450980392,0.827450980392,0.827450980392}
\definecolor{mpllightpink}{rgb}{1.0,0.713725490196,0.756862745098}
\definecolor{mpllightsalmon}{rgb}{1.0,0.627450980392,0.478431372549}
\definecolor{mpllightseagreen}{rgb}{0.125490196078,0.698039215686,0.666666666667}
\definecolor{mpllightskyblue}{rgb}{0.529411764706,0.807843137255,0.980392156863}
\definecolor{mpllightslategray}{rgb}{0.466666666667,0.533333333333,0.6}
\definecolor{mpllightslategrey}{rgb}{0.466666666667,0.533333333333,0.6}
\definecolor{mpllightsteelblue}{rgb}{0.690196078431,0.76862745098,0.870588235294}
\definecolor{mpllightyellow}{rgb}{1.0,1.0,0.878431372549}
\definecolor{mpllime}{rgb}{0.0,1.0,0.0}
\definecolor{mpllimegreen}{rgb}{0.196078431373,0.803921568627,0.196078431373}
\definecolor{mpllinen}{rgb}{0.980392156863,0.941176470588,0.901960784314}
\definecolor{mplmagenta}{rgb}{1.0,0.0,1.0}
\definecolor{mplmaroon}{rgb}{0.501960784314,0.0,0.0}
\definecolor{mplmediumaquamarine}{rgb}{0.4,0.803921568627,0.666666666667}
\definecolor{mplmediumblue}{rgb}{0.0,0.0,0.803921568627}
\definecolor{mplmediumorchid}{rgb}{0.729411764706,0.333333333333,0.827450980392}
\definecolor{mplmediumpurple}{rgb}{0.576470588235,0.439215686275,0.858823529412}
\definecolor{mplmediumseagreen}{rgb}{0.235294117647,0.701960784314,0.443137254902}
\definecolor{mplmediumslateblue}{rgb}{0.482352941176,0.407843137255,0.933333333333}
\definecolor{mplmediumspringgreen}{rgb}{0.0,0.980392156863,0.603921568627}
\definecolor{mplmediumturquoise}{rgb}{0.282352941176,0.819607843137,0.8}
\definecolor{mplmediumvioletred}{rgb}{0.780392156863,0.0823529411765,0.521568627451}
\definecolor{mplmidnightblue}{rgb}{0.0980392156863,0.0980392156863,0.439215686275}
\definecolor{mplmintcream}{rgb}{0.960784313725,1.0,0.980392156863}
\definecolor{mplmistyrose}{rgb}{1.0,0.894117647059,0.882352941176}
\definecolor{mplmoccasin}{rgb}{1.0,0.894117647059,0.709803921569}
\definecolor{mplnavajowhite}{rgb}{1.0,0.870588235294,0.678431372549}
\definecolor{mplnavy}{rgb}{0.0,0.0,0.501960784314}
\definecolor{mploldlace}{rgb}{0.992156862745,0.960784313725,0.901960784314}
\definecolor{mplolive}{rgb}{0.501960784314,0.501960784314,0.0}
\definecolor{mplolivedrab}{rgb}{0.419607843137,0.556862745098,0.137254901961}
\definecolor{mplorange}{rgb}{1.0,0.647058823529,0.0}
\definecolor{mplorangered}{rgb}{1.0,0.270588235294,0.0}
\definecolor{mplorchid}{rgb}{0.854901960784,0.439215686275,0.839215686275}
\definecolor{mplpalegoldenrod}{rgb}{0.933333333333,0.909803921569,0.666666666667}
\definecolor{mplpalegreen}{rgb}{0.596078431373,0.98431372549,0.596078431373}
\definecolor{mplpalevioletred}{rgb}{0.686274509804,0.933333333333,0.933333333333}
\definecolor{mplpapayawhip}{rgb}{1.0,0.937254901961,0.835294117647}
\definecolor{mplpeachpuff}{rgb}{1.0,0.854901960784,0.725490196078}
\definecolor{mplperu}{rgb}{0.803921568627,0.521568627451,0.247058823529}
\definecolor{mplpink}{rgb}{1.0,0.752941176471,0.796078431373}
\definecolor{mplplum}{rgb}{0.866666666667,0.627450980392,0.866666666667}
\definecolor{mplpowderblue}{rgb}{0.690196078431,0.878431372549,0.901960784314}
\definecolor{mplpurple}{rgb}{0.501960784314,0.0,0.501960784314}
\definecolor{mplred}{rgb}{1.0,0.0,0.0}
\definecolor{mplrosybrown}{rgb}{0.737254901961,0.560784313725,0.560784313725}
\definecolor{mplroyalblue}{rgb}{0.254901960784,0.411764705882,0.882352941176}
\definecolor{mplsaddlebrown}{rgb}{0.545098039216,0.270588235294,0.0745098039216}
\definecolor{mplsalmon}{rgb}{0.980392156863,0.501960784314,0.447058823529}
\definecolor{mplsandybrown}{rgb}{0.980392156863,0.643137254902,0.376470588235}
\definecolor{mplseagreen}{rgb}{0.180392156863,0.545098039216,0.341176470588}
\definecolor{mplseashell}{rgb}{1.0,0.960784313725,0.933333333333}
\definecolor{mplsienna}{rgb}{0.627450980392,0.321568627451,0.176470588235}
\definecolor{mplsilver}{rgb}{0.752941176471,0.752941176471,0.752941176471}
\definecolor{mplskyblue}{rgb}{0.529411764706,0.807843137255,0.921568627451}
\definecolor{mplslateblue}{rgb}{0.41568627451,0.352941176471,0.803921568627}
\definecolor{mplslategray}{rgb}{0.439215686275,0.501960784314,0.564705882353}
\definecolor{mplslategrey}{rgb}{0.439215686275,0.501960784314,0.564705882353}
\definecolor{mplsnow}{rgb}{1.0,0.980392156863,0.980392156863}
\definecolor{mplspringgreen}{rgb}{0.0,1.0,0.498039215686}
\definecolor{mplsteelblue}{rgb}{0.274509803922,0.509803921569,0.705882352941}
\definecolor{mpltan}{rgb}{0.823529411765,0.705882352941,0.549019607843}
\definecolor{mplteal}{rgb}{0.0,0.501960784314,0.501960784314}
\definecolor{mplthistle}{rgb}{0.847058823529,0.749019607843,0.847058823529}
\definecolor{mpltomato}{rgb}{1.0,0.388235294118,0.278431372549}
\definecolor{mplturquoise}{rgb}{0.250980392157,0.878431372549,0.81568627451}
\definecolor{mplviolet}{rgb}{0.933333333333,0.509803921569,0.933333333333}
\definecolor{mplwheat}{rgb}{0.960784313725,0.870588235294,0.701960784314}
\definecolor{mplwhite}{rgb}{1.0,1.0,1.0}
\definecolor{mplwhitesmoke}{rgb}{0.960784313725,0.960784313725,0.960784313725}
\definecolor{mplyellow}{rgb}{1.0,1.0,0.0}
\definecolor{mplyellowgreen}{rgb}{0.603921568627,0.803921568627,0.196078431373}

%% file: utils.tex
\newcommand{\RealismClassifier}[0]{{\textsc{RealismClassifier}}}
\newcommand{\Real}{\mathbb{R}}
\newcommand{\DCT}{\operatorname{DCT}}
\newcommand{\Tin}{{n}}
\newcommand{\Tout}{{N}}
\newcommand{\dObj}{{d_\text{Obj}}}
\newcommand{\pin}{x^{1:n}}
\newcommand{\pout}{x^{n+1:N}}

\newcommand{\cmark}[0]{\usym{2713}}

\newcommand{\mmark}[0]{$\thicksim$}

\newcommand{\csquare}[1]{\raisebox{0.25 em}{{\fcolorbox{#1}{#1}{\null}}}}

\newcommand{\ourOP}[0]{\operatorname{SAST}}

\newcommand{\ourUnder}[0]{Ours}

%% file: sections/1_abstract.tex
Forecasting long-term 3D human motion is challenging: the stochasticity of human behavior makes it hard to generate realistic human motion from the input sequence alone. Information on the scene environment and the motion of nearby people can greatly aid the generation process.
We propose a scene-aware social transformer model (SAST) to forecast long-term (10s) human motion motion. Unlike previous models, our approach can model interactions between both widely varying numbers of people and objects in a scene. We combine a temporal convolutional encoder-decoder architecture with a Transformer-based bottleneck that allows us to efficiently combine motion and scene information. We model the conditional motion distribution using denoising diffusion models.  We benchmark our approach on the Humans in Kitchens dataset, which contains 1 to 16 persons and 29 to 50 objects that are visible simultaneously. Our model outperforms other approaches in terms of realism and diversity on different metrics and in a user study. Code is available at \url{https://github.com/felixbmuller/SAST}.

\input{img/qualitative_sample_ours}

%% file: img/qualitative_sample_ours.tex
\begin{figure}[t]
    \centering

    \includegraphics[width=.5\textwidth, trim={8cm 8cm 8cm 12cm},clip]{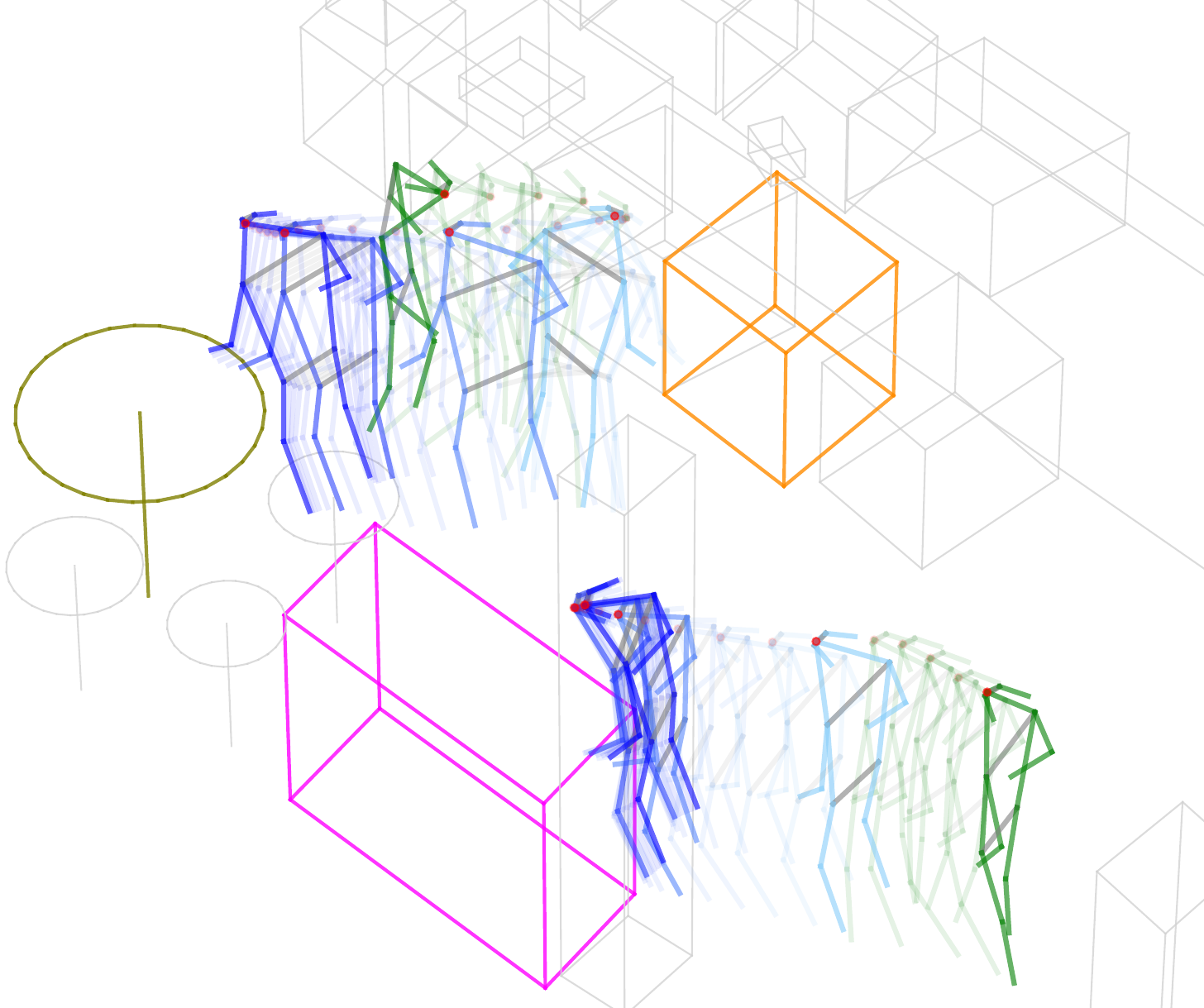}

    \caption{Our model forecasting complex realistic motion (fading from \csquare{mpllightskyblue} to \csquare{mplblue}) in a kitchen environment based on an input sequence \csquare{mplgreen} and scene context. The upper person returns to the standing table \csquare{mplolive} after grabbing something from the cupboard \csquare{mpldarkorange} while the lower person walks toward and stops at the large cupboard \csquare{mplmagenta}.}
    \label{fig:qualitative_sample_ours}
\end{figure}

%% file: sections/2_introduction.tex
Human motion forecasting aims to anticipate how humans may continue their movement in the future based on past observations \cite{fragkiadaki2015recurrent, jain2016structural, li2017auto, martinez2017human}. Humans perform this task instinctively to navigate complex multi-person environments \cite{schmidt1968anticipation}. It is also a highly relevant computer vision problem as it has various downstream applications, \eg  in robotics \cite{kantorovitch2014assistive, rosmann2017online}, healthcare \cite{kidzinski2020deep}, and neuroscience.

While short-term future motion is dominated by preservation of momentum, long-term future motion forecasting has to take the interdependence of motion in a multi-person setting into account. Humans interact with and react to each other all the time \cite{levinson2015timing, sacks1974simplest}. The environment is also of crucial importance  for motion forecasting \cite{DBLP:conf/iccv/AdeliE0NS0R21}, as it predetermines the space of possible behaviors. The placement of \eg walkways, doors, and chairs greatly impacts what behaviors can be expected in certain situations.

While long-term motion forecasting has seen a proliferation of work in recent years \cite{Ng_2024_CVPR, Mughal_2024_CVPR, jeong2024multi, Diller_2024_CVPR, barquero2024seamless, fan2024freemotion, liang2024intergen, guo2022generating, li2024two, DBLP:conf/iccv/TankeZZTCWWGK23}, most approaches still focus on the single  or two-person special case. Especially multi-person situations with a fluctuating number of people are understudied despite their common occurrence in the wild. Also, few works consider the scene environment despite its rich information content. TriPod \cite{DBLP:conf/iccv/AdeliE0NS0R21} aims to address this by 
employing two graph attention modules for human-object and human-human interactions with message passing in between. However, TriPod does not scale well to sequences longer than a second, presumably due to its autoregressive approach using RNNs.

We want work towards unconstrained human motion models that are able to model realistic interactions in the wild and we aim to bridge this gap with our scene-aware social
transformer model (SAST) for long-term forecasting. 
Our model allows for

\begin{itemize}
    \item long-term (10 seconds) motion forecasting,
    \item versatile interaction modeling for widely varying (1-16) numbers of persons,
    \item scene-agnostic environment modeling based only on variable numbers (50+) of 3D object point clouds, and
    \item sampling of multiple realistic continuations from the conditional motion distribution.
\end{itemize}

Jointly forecasting multi-person motion for a large varying number of persons is difficult to learn. We simplify this task by forecasting only one person at a time during training (with context information from other people). During inference, we are able to produce highly interdependent multi-person motion by exchanging motion information throughout the diffusion process.

To our knowledge, our approach is the first long-term multi-person motion forecasting model that takes scene context into account. To showcase our flexible modeling approach, we evaluate our model on the Humans in Kitchens dataset \cite{DBLP:conf/nips/TankeKMDG23} with 1 to 16 persons interacting simultaneously in four different environments containing 37 to 50 objects, predicting up to 10 seconds of output motion.

%% file: sections/2.5_related_work.tex
Human pose forecasting aims to predict a future pose sequence given some pose history \cite{li2017auto, martinez2017human, jain2016structural, fragkiadaki2015recurrent}. There has been considerable work on single-person forecasting \cite{DBLP:conf/eccv/MaoLS20, DBLP:conf/iccv/MaoLSL19, DBLP:conf/3dim/AksanKCH21, DBLP:conf/eccv/LucasBWR22, butepage2017deep, holden2015learning, li2018convolutional, guo2023back}. As multi-person interactions play a crucial role in forming human behavior, significant work has also been done on modeling human-human-interactions \cite{DBLP:conf/iccv/AdeliE0NS0R21, DBLP:journals/corr/abs-2208-09224, DBLP:conf/nips/VaswaniSPUJGKP17, DBLP:conf/nips/WangXNW21, DBLP:journals/corr/abs-2208-14023}. However, even in the multi-person case, most prior work focuses on two-person \cite{Ng_2024_CVPR, Mughal_2024_CVPR, liang2024intergen, DBLP:conf/iccv/AdeliE0NS0R21} or three-person \cite{DBLP:conf/nips/WangXNW21, DBLP:conf/iccv/TankeZZTCWWGK23} scenarios.

\paragraph{Long-Term Forecasting} Initially, motion forecasting focused on short-term forecasting up to 1 second \cite{li2017auto, martinez2017human, jain2016structural, fragkiadaki2015recurrent, guo2023back, DBLP:conf/eccv/MaoLS20, DBLP:conf/iccv/AdeliE0NS0R21}. Diffusion probabilistic models \cite{DBLP:conf/nips/HoJA20} sparked the development of both single-person and multi-person models forecasting several seconds of realistic motion \cite{DBLP:conf/iccv/TankeZZTCWWGK23, Ng_2024_CVPR, Mughal_2024_CVPR, barquero2024seamless, liang2024intergen}, even though non-diffusion long-term-forecasting approaches have been proposed as well \cite{jeong2024multi, Diller_2024_CVPR}. Multi-second forecasting horizons are necessary to allow for diverse downstream application \cite{jeong2024multi}. But in the area of long-term forecasting, the stochastic nature of human behavior  requires turning away from exact ground truth reconstruction toward modeling the distribution of plausible motion sequences \cite{barquero2023belfusion} given the input sequence and additional context if available.
 
\paragraph{Motion Synthesis with Context Information} There has been some work on combining long-term motion forecasting with guiding signals like speech \cite{Ng_2024_CVPR, Mughal_2024_CVPR} or action labels \cite{barquero2024seamless, fan2024freemotion, liang2024intergen}. However, few works take scene context into account, even though it contains crucial information for many human behaviors and is present in many in-the-wild scenarios \cite{DBLP:conf/nips/TankeKMDG23, Diller_2024_CVPR}. Existing work employing scene context is either limited to short-term \cite{DBLP:conf/iccv/AdeliE0NS0R21} or single-person \cite{Diller_2024_CVPR} forecasting.

%% file: sections/3_methodology.tex
\subsection{Problem Representation}

\paragraph{Motion Sequences} Given $P$ persons with $J$ joints, the motion sequence of person $i$ consisting of $N$ frames is 
\begin{align}
    X^{(i),1:N} &= [j_{1, 1}, j_{1, 2}, \dots, j_{J, 3}] \text{ with } j_{k,d} \in \Real^N
\end{align}
where each joint $j^{i}_{k, d}$ is in global Cartesian coordinates. We denote the concurrent motion sequences of $P$ persons as 
\begin{equation}
    X = \left[X^{(1)}, X^{(2)}, \dots, X^{(P)}\right] \in \Real^{P \times J \times 3 \times N}.
\end{equation}

\paragraph{Scene Geometry}

We describe the scene geometry (\eg walls, tables, and chairs) as a set of point clouds whose positions may change  over time, for example, by moving a chair. We use basis point set (BPS) encoding \cite{DBLP:conf/iccv/ProkudinL019}  to represent each object at a specific point in time by a fixed-size vector. BPS encoding is well suited to provide compact representations for objects of widely varying sizes and shapes. It represents objects by the distance between random fixed basis points and the closest point on the object surface. The number of basis points equals the length of the resulting vector representation and can be chosen to balance between granularity and encoding size. We also concatenate a one-hot encoding of 13 object types to the BPS encoding, \eg chair, table, or coffee machine. The encoding dimensionality is $\dObj{} = 2,061$. We define a scene with $G$ objects at time $n$ as

\begin{equation}
    S^{n} = \left\{S^{(1),n}, S^{(2),n}, \dots, S^{(G),n}\right\} \text{ with } S^{(i),n} \in \Real^\dObj{}
\end{equation}

\paragraph{Forecasting Objective}

Our goal is to learn a social interaction model that is able to generate a realistic and plausible future multi-person motion sequence given pose history and the scene. $\Tin$ denotes the number of frames in the input sequence, while $\Tout$ is the number of frames in the whole sequence. We only predict future motion for people present in the input sequence $X^{1:\Tin}$, even if more people enter the scene during the ground truth output sequence. We use zero-velocity padding if a person enters late during the input sequence or leaves early during the ground truth output sequence. In this work, we limit our model to static scene geometry, i.e. the state of the scene at the last input frame.

\begin{equation}
    \hat{X}^{\Tin+1:\Tout} = \ourOP(X^{1:\Tin}, S^{\Tin}) 
\end{equation}

\subsection{Normalization and Scaling}

We design our model to forecast the motion of one person at a time (the primary person) given environment context (other persons and scene). We align the primary pose sequences $X^{(i)}$  at the last frame of the input sequence $\Tin$ such that the mean of both hip joints is at $(x, y) = (0, 0)$ and the hip is parallel to the x-axis. This removes global translation and rotation from the primary pose sequence. We do not perform normalization on the z-axis, as all motion in the dataset we are working with is based on an equal-level floor. This normalization criteria induce a affine transformation $\operatorname{Norm}_i$, which we also apply to the environment context, similar to \cite{jeong2024multi}. Given one multi-person motion sequence $X$ and a scene context $S$, we thus create $P$ normalized data points $\{(x^{(1)}, O^{(1)}, s^{(1)}), \dots, (x^{(P)}, O^{(P)}, s^{(P)})\}$ with $x$ being the respective primary pose sequence, $O$ being the context of other persons, and $s$ being the scene context.
\begin{alignat}{2}
\label{eg:norm}
    x^{(i)} &= \operatorname{Norm}_i(X^{(i)}) &&\in \Real^{J \times 3 \times N}\\
    O^{(i)} &= \left[\operatorname{Norm}_i(X^{(j)}) \;\big|\; \forall j \ne i\right] &&\in \Real^{(P-1)\times J \times 3 \times N} \nonumber \\
    s^{(i)} &= \operatorname{Norm_i}(S_n) &&\in \Real^{G\times \dObj} \nonumber
\end{alignat}

During the diffusion process, pose sequences are disturbed using noise from a normal distribution. To achieve distributional similarity between the data and noise distribution, we perform min-max-scaling of $x$ to $[-3, 3]$. This scaling maps all motion sequences in the training set to be with $3\sigma$ of a unit Gaussian thus enabling the model to generate all motion present in the training data. We find min-max-scaling to produce superior qualitative results compared to the commonly used Normal scaling.

\subsection{Diffusion Model}

We model the generation of realistic motion sequences as denoising task using diffusion models. Our model uses latent variables $x_1, \dots, x_T$ of equal dimensionality with $x_0 \sim q(x_0)$ being a noise-free single-person motion sequence. The diffusion process $q$ is a Markov chain specifying $x_1, \dots, x_T$ as progressively more noisy versions of $x_0$. This can be expressed as closed formula 
\begin{equation}
\label{eq:diffusion_forward}
    q(x_t | x_0) = \mathcal{N}(x_t; \sqrt{\Bar{a}_t}x_0, (1- \Bar{a}_t) \mathbf{I})
\end{equation}
given some variance schedule $\beta_t$ with $\alpha_t = 1-\beta_t$ and $\Bar{a}_t = \prod_{s=1}^t \alpha_s$. The variance schedule $\beta_t$ is chosen such that $x_T \sim \mathcal{N}(0, \mathbf{I})$ approximately holds. 

The reverse process $p_\theta$ consists of learned denoising transitions $p_\theta(x_{t-1}|x_t)$. Instead of directly learning $p_\theta$, we learn a function $f_\theta(x_t, C, t)$ predicting the noise-free motion sequence $\hat{x}_0$.
To condition the model on the input sequence and context, we pass a context tuple 
\begin{equation}
    C = (\pin, O, s)
\end{equation}
to $f_\theta$ consisting of the noise-free input sequence $\pin$, other person trajectories $O$ and scene context $s$. Given a predicted $\hat{x}_0$, we calculate the reverse step as 
\begin{align}
\label{eq:reverse_step}
    p_\theta(x_{t-1}|x_t) &= \mathcal{N}(x_{t-1}; \Tilde{\mu}_t(x_t, \hat{x}_0), \Tilde{\beta}_t \mathbf{I}) \text{ with }\\
    \Tilde{\mu}_t(x_t, \hat{x}_0) &= \frac{\sqrt{\Bar{\alpha}_{t-1}}\beta_t}{1 - \Bar{\alpha}_t} \hat{x}_0 + \frac{\sqrt{\alpha_t}(1-\Bar{\alpha}_{t-1})}{1-\Bar{\alpha}_t} x_t \nonumber \text{ and }\\
     \Tilde{\beta}_t &= \frac{1- \Bar{\alpha}_{t-1}}{1- \Bar{\alpha}_t} \beta_t . \nonumber
\end{align}

\paragraph{Training}

Given a training datapoint $(x, O, s)$, we use Equation~\ref{eq:diffusion_forward} to calculate the loss of our denoising model $f_\theta$ as
\begin{equation}
    L = \Vert \pout - f_\theta(\sqrt{\Bar{a}_t}x + \sqrt{1-\Bar{a}_t} \epsilon, C, t)^{\Tin+1:\Tout} \Vert
\end{equation}
with random noise $\epsilon \sim \mathcal{N}(0, \mathbf{I})$ and timestep $t \sim \text{Uniform}(1 \dots T)$. We exclude padded frames (due to people entering or leaving the scene) from the loss calculation. We use $L^1$ loss, as it is more robust against outliers compared to $L^2$. This is relevant for human motion datasets, as joint positions are usually estimated with motion capture algorithms \cite{DBLP:conf/nips/TankeKMDG23}, introducing some noise in the joint locations. 

\paragraph{Joint Multi-Person Inference}

To forecast a multi-person motion sequence, we sample $x_T \sim \mathcal{N}(0, \mathbf{I})$ and iteratively denoise it using $f_\theta$ and Equation~\ref{eq:reverse_step}. As we only see $X^{1:n}$ during inference, only $O^{1:\Tin}$ is known while $O^{\Tin+1:\Tout}$ is unknown. But to allow for interdependent motion generation, it is crucial that the calculation of $f_\theta(x^{(i)}_t, C^{(i)}, t)$ for the primary person $i$ sees the whole motion sequences of all other people $O^{(i),1:\Tout}$. To allow this important information flow between different persons, we perform all inference diffusion processes for $X$ in parallel and de-normalize the predicted noise-free sequences $\hat{x}_{0,t-1}$ of all people from diffusion step $t-1$ as
\begin{equation}
    \hat{X}_{t-1} = \left[\operatorname{Norm}^{-1}_i(\hat{x}_{0,t-1}^{(i)}) \mid \forall i \in \{1, \dots, P\}\right].
\end{equation}
We then estimate $O^{1:N}_t$ for every person $i$ based on $\hat{X}_{t-1}$ using Equation~\ref{eg:norm}. This $O_t^{(i), 1:N}$ is then used in the next inference step $t$. We bootstrap this process using zero-velocity padding of $O^{1:\Tin}$ for $t=0$. This approach provides an iteratively improving estimation of the behavior of other people in the scene throughout the diffusion inference.

\input{img/model_overview}

\subsection{Denoising Model}

\newcommand{\Enc}{{\text{Enc}}}
\newcommand{\Dec}{{\text{Dec}}}
\newcommand{\Mix}{{\text{Mix}}}

The UNet-style \cite{DBLP:conf/miccai/RonnebergerFB15} denoising model 
\begin{equation}
    \hat{x}_0^{1:N} = f_\theta(x^{1:N}_t, (\pin, O, s), t)
\end{equation}
consists of a temporal convolutional encoder $e$ and a temporal convolutional decoder $d$, as well as a Transformer-based aggregation module, which combines pose and scene information in the bottleneck between encoder and decoder. See Figure~\ref{fig:model_overview} for an overview.

\paragraph{Pose Encoder}

Following \cite{DBLP:journals/corr/abs-1803-01271}, we let $e$ be a three-layer temporal convolutional network, with each layer consisting of two causal convolutional submodules with residual connections. The diffusion step $t$ is encoded using Gaussian Fourier projection and fed into each convolutional layer. In each layer, we use strided convolution to halve the temporal resolution.

Two encoder $e^x$ and $e^O$ are applied to the primary and other pose sequences respectively. We add the noise-free input sequence $\pin$ to the noisy motion sequence $x^{1:N}_t$ by performing zero-velocity padding to $1:N$ and concatenating on the body joint dimension. We use skip connections between each layer of $e^p$ and each layer of $d$.
\begin{alignat}{2}
    \text{skip}, h_x &= e^x(\pin \Vert \pout_t, t) &&\in \Real^{N \times D} \\
    h_O &= [e^O(O^{(1)}, t) \Vert \dots \Vert e^O(O^{(P-1)}, t)] &&\in \Real^{((P-1) \times N) \times D} \nonumber
\end{alignat}

\paragraph{Aggregation Module}

The aggregation module consists of two Transformer \cite{DBLP:conf/nips/VaswaniSPUJGKP17} modules combining $h_x$, $h_O$, and $s$. 
\begin{align}
    h &= \text{TDec}^s(h_x, \text{TEnc}^s(s)) \\
    h' &= \text{TDec}^O(h, \text{TEnc}^O(h_O)) \nonumber
\end{align}

$\text{TDec}^s$, $\text{TDec}^O$, and $\text{TEnc}^O(h_O)$ use sinusoidal positional encodings \cite{DBLP:conf/nips/VaswaniSPUJGKP17} for encoding the order of motion tokens and attention masking to ensure causality.

\paragraph{Pose Decoder}

The decoder $d$ is a three-layer temporal convolutional network. The diffusion step is supplied via Gaussian Fourier Projection and we use linear upsampling in each layer to double the temporal resolution. Skip connections from $e^x$ are incorporated in each layer. The noise-free output sequence is calculated as 

\begin{equation}
    \hat{x}_0 = d(h', \text{skip}, t).
\end{equation}

%% file: img/model_overview.tex
\begin{figure}[t]
    \centering
    \includegraphics[width=\linewidth]{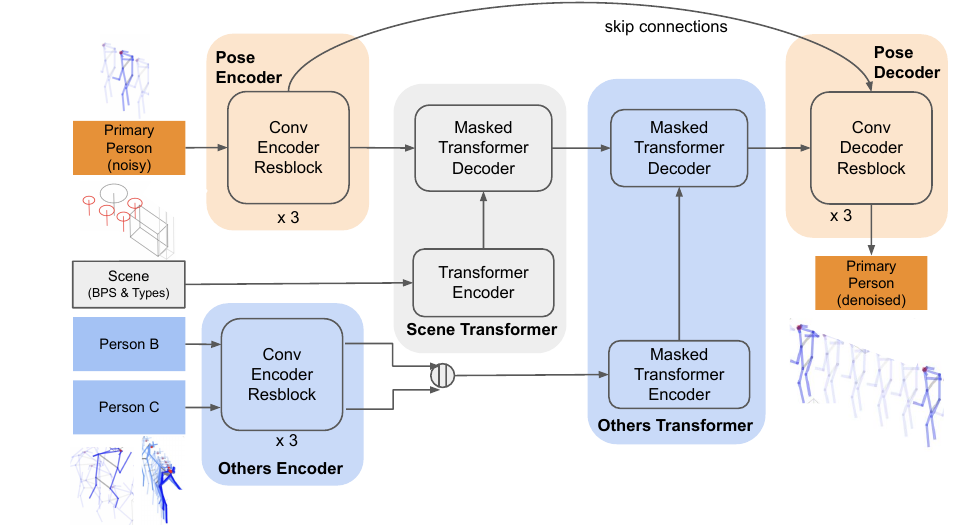}
    \caption{Architecture of the denoising model $f_\theta$.}
    \label{fig:model_overview}
\end{figure}

%% file: sections/4_experiments.tex
\paragraph{Metrics}

We follow evaluation procedures established in prior work on long-term motion forecasting \cite{DBLP:conf/nips/WangXNW21, tanke2021intention, DBLP:conf/iccv/TankeZZTCWWGK23, Diller_2024_CVPR}. To measure local realism, \ie realism on short motion snippets, we use the velocity-based \emph{NDMS} \cite{tanke2021intention} scores and realism scores by a model trained to discriminate between real and synthetic motion samples \cite{Diller_2024_CVPR}. We also use the \emph{distribution of trajectory lengths} as an auxiliary measure for global motion realism \cite{DBLP:conf/nips/WangXNW21}, \ie overall realism of the whole predicted sequence. We also perform a user study.

In addition to those metrics, we propose to use \emph{UMWR}, a local metric for motion diversity complementing NDMS, and  \emph{mean velocity over time} as another auxiliary measure for global realism.

\paragraph{Dataset} Humans in Kitchens \cite{DBLP:conf/nips/TankeKMDG23} is a 3D motion capture dataset of people interacting in a coffee kitchen environment with minimal instructions. The recording time is 7h over four recordings with 90 unique persons in total and 1 to 16 persons visible simultaneously. Each of the four recordings takes place in a different coffee kitchen, hence having a different scene geometry with 29 to 50 objects. We follow the evaluation protocol proposed by Tanke \etal \cite{DBLP:conf/nips/TankeKMDG23}, using the recordings A-C as training data and evaluating on a subset of sequences from D. The Humans in Kitchens evaluation set focuses on transitional moments. Sequences were selected so that the end of the input sequence marks the beginning of a new action (sitting down, standing up, opening the fridge, etc.), which a model is supposed to predict. The evaluation setup requires models to generalize to a different scene geometry than seen during training, containing both more objects (train: 40, test: 50) and more simultaneously visible persons (train: 14, test: 16).

\paragraph{Baselines} We follow \cite{DBLP:conf/nips/TankeKMDG23} in their selection of baselines. MRT \cite{DBLP:conf/nips/WangXNW21} is a purely Transformer-based architecture for multi-person motion forecasting, while SiMLPe \cite{guo2023back} and HisRep \cite{DBLP:conf/eccv/MaoLS20} are single-person models. TriPod \cite{DBLP:conf/iccv/AdeliE0NS0R21} is a graph-attention-based model for multi-person motion forecasting with scene context.

\subsection{Model Training}

We use Adam with weight decay \cite{DBLP:conf/iclr/LoshchilovH19} with a linear learning rate schedule from \num{2e-7} to \num{5e-5}. We train for 680k training steps with a batch size of 32. For the diffusion process, we use $T = 1,000$ steps and the cosine variance schedule \cite{DBLP:conf/icml/NicholD21}.  As the pose types in Humans in Kitchens are heavily imbalanced, we undersample motion sequences that contain only standing poses by 50\%. This removes 15\% of the training data.

For the Scene Transformer, we use 3 encoder and decoder layers with $d_\text{Scene}=256, d_\text{ff} = 1024,$ and 8 heads. For the Others Transformer, we use 2 encoder and decoder layers with $d_\text{Others}=128, d_\text{ff} = 512,$ and 4 heads, since its input is preprocessed with the convolutional Others Encoder. Our model has 15.3M parameters.

\input{img/realism_classifier_schema}

\input{tables/local_metrics}

\subsection{Local Realism and Diversity}

\paragraph{Classifier-based Realism Metric}

To judge the local realism of generated motion, we train an evaluation model to distinguish real and synthetic motion. We train on 50\% real samples and 10\% predicted sequences from SiMLPe, MRT, TriPod, HisRep, and Ours each. As the classifier is equally well trained to detect synthetic motion for each model, we can assume that a higher mean realism score is indicative of a model producing more realistic motion than other models. 

The classifier has a simple feed-forward architecture, see Figure~\ref{fig:realism_classifier}, and works on an input sequence length of 50 frames. See the supplementary material for implementation details. The classifier has excellent classification performance. On the validation set (92,997 samples) it achieves an accuracy of $0.997$ and an Area under the ROC curve of $0.9994$, i.e. the model ranks a random real sample higher than a random synthetic sample with 99.94\% probability.

\paragraph{NDMS}  Normalized Directional Motion Similarity \cite{tanke2021intention, DBLP:conf/iccv/TankeZZTCWWGK23} employs a set of short real motion snippets as a reference set $\mathcal{D}$ for realistic motion. To evaluate a prediction $\hat{X}^{1:\Tout}$, it is split into short motion snippets $x$. Each $x$ is matched to the most similar $\Tilde{x} \in \mathcal{D}$ and a velocity-based score on $x$ and $\Tilde{x}$ is calculated. 

Following \cite{tanke2021intention}, we prepend each predicted sequence with a few frames of its input sequence to also measure the realism of the transition between input and predicted sequence. We build the reference set $\mathcal{D}$ from the test split.  We report the mean NDMS score over all frames in all predicted sequences.

\paragraph{UMWR} We propose using the reference set $\mathcal{D}$ of NDMS for judging diversity as well. The \emph{unique motion word ratio} is calculated as

\newcommand{\opNN}{\operatorname{NN}}

\begin{equation}
    \operatorname{UMWR}(\chi) = \frac{\big|\left\{\opNN(\chi^{1:\kappa}), \opNN(\chi^{2:\kappa+1}), \dots \opNN(\chi^{|\chi|+1-\kappa:|\chi|})\right\}\big|}{|\chi|+1-\kappa}
\end{equation}
where $\kappa = 8$ frames is the motion snippet length, $\chi = X^{(i),\Tin+2-\kappa:\Tout}$ is the single-person sequence we want to evaluate (output sequence and the last frames of the input sequence), and $\opNN(x)$ is the function to map $x$ to the most similar $\Tilde{x} \in \mathcal{D}$.

UMWR scores are in $[0, 1]$, with a low UMWR implying that the motion is less diverse as most of the generated motion is within a very small region in the possible motion space, namely close to a few motion words in $\mathcal{D}$. To calculate UMWR on subsequences, we use $\operatorname{UMWR}@k\operatorname{s} = \operatorname{UMWR}(\chi^{25(k-1):25k})$ as the frame rate is 25 frames per second.

\paragraph{Results} 

Table~\ref{tab:local_metrics} contains the local realism and diversity results. We observe that our model performs best for sequences of six seconds or longer. SiMLPe performs very good for short sequences, but performance rapidly drops for longer sequences. On NDMS, our model scores worse than SiMLPe and HisRep. In terms of diversity, our model performs best. The UMWR score decreases after the first seconds, but it is consistently higher than for other baseline models.

\input{img/trajectories}

\input{tables/user_study}

\subsection{Global Realism and Diversity}

To judge the whether models create realistic global behaviour, we compare the distribution of trajectories generated by models to the distribution of trajectories in the ground truth dataset. For each motion sequence $X = [j_{1,1}, j_{1, 2}, \dots, j_{J, 3}]$, we define the 2D root trajectory $r$ as
\begin{equation}
    r = \begin{bmatrix}
        (j_{13,1} + j_{14,1}) / 2 \\
        (j_{13,2} + j_{14,2}) / 2 \\
    \end{bmatrix} \in \Real^{N \times 2}
\end{equation}
with $j_{13}$ and $j_{14}$ being the left and right hip joint.

\input{img/velocity_over_time}

For qualitative judgement, we visualize a random subset of trajectories, see Figure~\ref{fig:trajectories}. We find that our model produces trajectories that are visually similar to the ground truth. Realistic trajectories seem to follow a random-walk like pattern with varying trajectory lengths. TriPod does not produce short trajectories, only long and regular ones. This is caused by TriPod generating a constant shifting motion, see the supplemantary material for example videos. We also visualize the velocity over time for all models (mean over all test samples) in Figure~\ref{fig:velocity_over_time}. TriPod matches the ground truth mean velocity most closely, however this is due to constant shifting motion in its predictions. Our model produces faster-than-realistic motion in the first three seconds and produces slower-than-realistic motion afterwards. Other baselines produce hardly any motion after two seconds.

A direct quantitative analysis of the trajectory distribution is not feasible because of the high dimensionality and limited size of the test set. Instead we analyze the distribution of trajectory length

\begin{equation}
    d = \frac{1}{N-n} \sum_{t=n+1}^N \parallel r^t - r^{t-1} \parallel_2 .
\end{equation}

Let $\mu$ be the probability distributions over trajectory lengths for the ground truth and let $\nu_M$ be the distribution for a model $M$. We sample from $\mu$ and $\nu_M$ by calculating the trajectory length $d$ for all output sequences in the test set (both ground truth and generated by $M$). Using this samples, we estimate the Wasserstein distance between $\mu$ and $\nu_M$, see Table~\ref{tab:distance_moved_wasserstein}. We find that our model resembles the ground truth distribution the most.

\subsection{User Study and Qualitative Evaluation}

Covering all aspects of realistic motion with metrics is tricky and prior work has repeatedly reported discrepancies between metrics and subjective evaluation by users \cite{Mughal_2024_CVPR, dabral2023mofusion, tseng2023edge}. We therefore perform a user study to judge the realism of generated human motion. In the user study, we let humans rank model predictions as well as ground truth sequences from most to least realistic. Model names were blinded and we collected 43 rankings from 10 persons. Samples used in the user study and detailed results can be found in the supplementary material. We report the results of the user study in Table~\ref{tab:user_study}. Our model is ranked highest, followed by SiMLPe. 
 
This result is consistent with our qualitative observations. SiMLPe is able to produce the longest smooth continuation of the input sequence among the baselines, but freezes in an average pose after around two seconds. MRT sometimes generates realistic global motion, but mostly freezes similar to SiMLPe and HisRep. TriPod produces an unrealistic slow drifting motion for all people in the scene.

\input{img/viz_diversity}

Our model is able to produce a wide variety of realistic motion, like sitting down, standing up, or directed movement towards an object in the scene. It also produces realistic object interactions with \eg cupboards, fridges, or whiteboards. 
Prompting our model multiple times generates multiple realistic samples from $ p(X^{\Tin+1:\Tout} | X^{1:\Tin}, S^{\Tin})$. Figure~\ref{fig:viz_diversity} shows a qualitative example.

But qualitative evaluation also shows some weaknesses of our model: It produces more high-velocity motion in the first frames of the predicted sequence than the ground truth would suggest, sometimes causing visible discontinuities to the input sequence. Motion sequences with long global motion also tend to have less realistic local limb movement, \eg missing leg movement.

\input{tables/ablations}

\subsection{Ablations}
\label{sec:ablations}

To study the impact of multi-person and scene context, we ablate the other person encoder and the scene encoder respectively. We ablate a module by replacing its output with constant all-zero vectors during training and inference. This produces attention weights of zero and thus effectively ignores the module in the decoder. 

We provide a quantitative analysis in Table~\ref{tab:ablations}. As none of the existing metrics are able to explicitly judge human-human or human-scene interactions, we also provide a qualitative analysis here. Samples of all models are provided in the supplementary material.

\paragraph{Scene Context}
Without scene context, the quality of human-scene interactions is reduced. The model does not produce directed motion with a clear goal anymore, \ie walking to a cupboard and stopping there. While the model is still able to produce sensible standing up and sitting down motion, sitting down without a chair beneath a person happens more frequently. This is plausible, as the the beginning of a sitting down or standing up motion is usually apparent at the end of the input sequence, but precise scene information is missing. The evaluation set of Humans in Kitchens is explicitly designed to focus on transitional moments based on action labels.

\paragraph{Other People}
We do not see a deterioration of motion realism when ablating the other person encoder. However, without the other person encoder, our model is unable to produce synchronized multi-person motion anymore. Our model sometimes predicts that 2-3 persons sitting next to each other get up even if the start of the standing up motion is only apparent in the input sequence of one person, which does not happen when the other person encoder is ablated. This shows that the joint inference procedure we are using allows to model interdependent motion and that the information flow between multiple persons has a precise temporal and spatial resolution.

%% file: img/realism_classifier_schema.tex
\begin{figure*}
    \centering
    \includegraphics[width=0.9\textwidth]{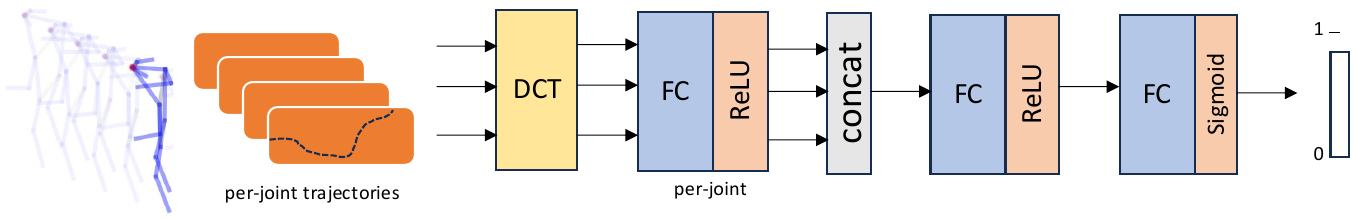}

    \caption{The realism scoring model calculates a score based on short single-person motion snippets. It is trained to distinguish real and synthetic motion, the latter is generated using our model and the baseline models.}
    \label{fig:realism_classifier}
\end{figure*}

%% file: tables/local_metrics.tex
{
\setlength{\tabcolsep}{3pt}

\begin{table}
    \centering
\caption{Local realism (Realism Score, NDMS) and diversity (UMWR) metrics. Realism score at $k$ is the mean realism score on $[0,k)$.}
    \begin{tabular}{lrrrrrcrrrrr}
    \toprule
           &   \multicolumn{5}{c}{Realism Score $\times 100$ $\uparrow$} & \multirow{2}{*}{NDMS $\uparrow$}& \multicolumn{5}{c}{UMWR $\uparrow$} \\
         \cline{2-6} 
         \cline{8-12}
          & 2s  & 4s & 6s  & 8s  & 10s && 2s & 4s & 6s & 8s & 10s \\
          \midrule
         
         MRT  & 4.23  & 1.76  & 1.10  & 0.85  & 0.77 & 0.16 & 0.09 & 0.08 & 0.07 & 0.07 & - \\
         HisRep  & \underline{8.35}  & 1.10  & 0.58  & 0.40  & 0.30 & \underline{0.23} & 0.12 & 0.07 & 0.06 & 0.06 & 0.06 \\
         SiMLPe & \textbf{18.24}  & \textbf{4.16}  & \underline{2.40} &  \underline{1.71}  & \underline{1.33} & \textbf{0.27} & 0.15 & 0.07 & 0.07 & 0.06 & 0.06 \\
         TriPod  & 3.39  & 0.37  & 0.22  & 0.17  & 0.14 & 0.13 & \underline{0.18} & \underline{0.14} & \underline{0.12} & \underline{0.11} & \underline{0.11} \\
         \midrule
         Ours  & 5.75  & \underline{2.80} &  \textbf{2.88}  & \textbf{2.55} & \textbf{2.40} & 0.17 & \textbf{0.41} & \textbf{0.21} & \textbf{0.15} & \textbf{0.14} & \textbf{0.15} \\
         \bottomrule
    \end{tabular}

    \label{tab:local_metrics}
\end{table}

}

%% file: img/trajectories.tex
\begin{figure*}[t]

    \makebox[\textwidth][c]{\includegraphics[width=1.05\textwidth,clip]{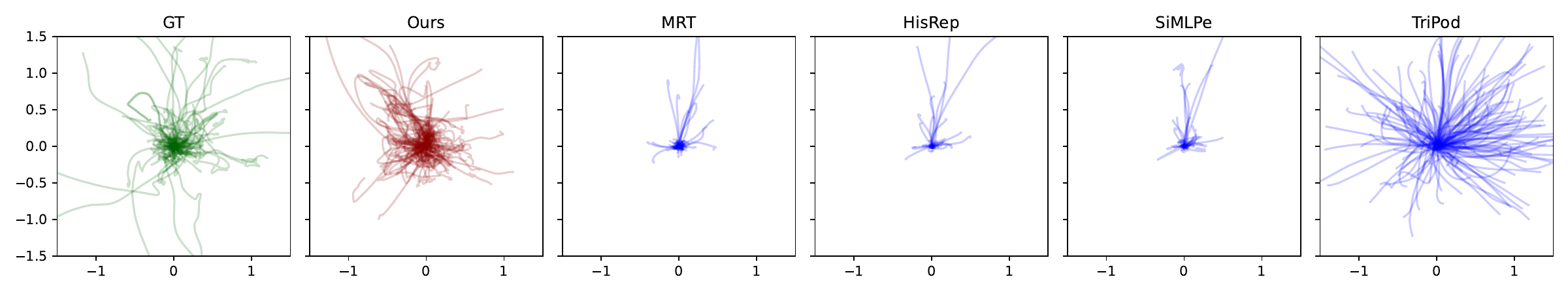}}

    \caption{Visualization of ten-second output trajectories $r^{n:N}$ for each model. The last frame of the input sequences is normalized to $(0,0)$ with persons facing in positive $y$-direction. 20 randomly selected trajectories per model displayed.}
    \label{fig:trajectories}
\end{figure*}

%% file: tables/user_study.tex
\begin{table}[]
    \centering
\parbox{.47\linewidth}{
    \centering
    \caption{Comparison between our models and four baselines in the user ranking. Users were asked to rank all models and the ground truth from most (1) to least (6) realistic and we provide the mean and standard deviation based on 43 rankings. See the supplementary material for details.}
    \begin{tabular}{lrrr}
    \toprule
    &  \multicolumn{2}{c}{User Ranking} \\
    \cmidrule(lr){2-3}
    & Mean $\downarrow$  & Std  \\
    \midrule
    Ground Truth & 1.30 & 0.74  \\
    \midrule
    MRT \hfill \cite{DBLP:conf/nips/WangXNW21}  & 4.44 & 1.01  \\
    HisRep \hfill\cite{DBLP:conf/eccv/MaoLS20} & 4.98 & 0.96  \\
    SiMLPe \hfill\cite{guo2023back} & \underline{3.62}  & 1.25 \\
    TriPod\hfill\cite{DBLP:conf/iccv/AdeliE0NS0R21} & 4.37 & 1.48  \\
    \midrule
    Ours & \textbf{2.23} & {1.02}  \\
    \bottomrule
    \end{tabular}
    
    \label{tab:user_study}
    }
    \hfill
    \parbox{.47\linewidth}{
    \centering
    \caption{Comparison of the distribution of trajectory lengths between ground truth and models. For each distribution, we give the mean, standard deviation, and Wasserstein distance $W_1$ to the ground truth distribution. Best Wasserstein distance and closest mean to the ground truth is highlighted.}
    \begin{tabular}{lrrr}
        \toprule
        & \multicolumn{3}{c}{Trajectory Dist.} \\
        \cmidrule(lr){2-4}
        & Mean & Std & $W_1$ $\downarrow$ \\
        \midrule
        Ground Truth & 1.17 & 1.70 & - \\
        \midrule
        MRT & 0.20 & 0.36 & 0.96 \\
        HisRep & 0.26 & 0.44 & 0.91\\
        SiMLPe & 0.26 & 0.39 & 0.91\\
        TriPod & \textbf{1.16} & 0.54 & \underline{0.72}\\
        \midrule
        Ours & \underline{0.86} & 0.62 & \textbf{0.57}\\ 
        \bottomrule
    \end{tabular}
    
    \label{tab:distance_moved_wasserstein}
    }
    
\end{table}

%% file: img/velocity_over_time.tex
\begin{figure}[t]
    \centering
    \includegraphics[width=0.9\linewidth, trim={0 0.3cm 0 0}, clip]{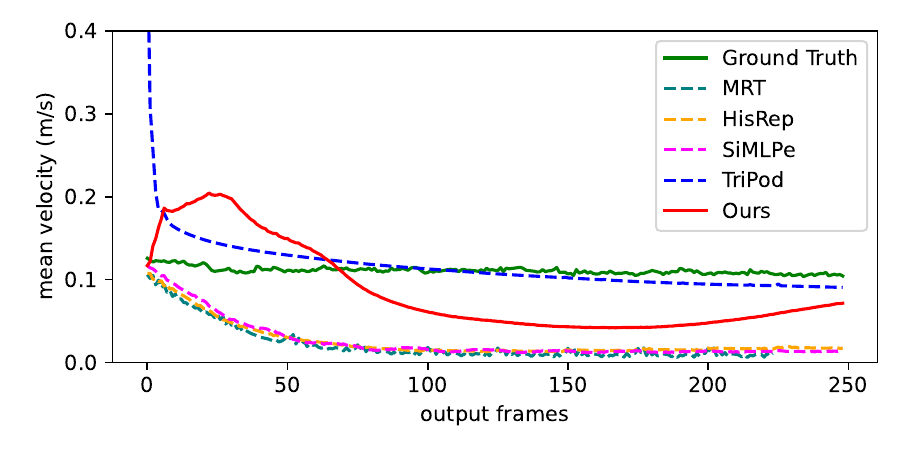}
    \caption{Frame-wise mean global velocity for all models. We calculate the velocity of the hip center in the x- and y-direction and average over all evaluation samples. Outliers in the ground truth data (few single-frame velocities over 10 m/s) are clipped before averaging. }
    \label{fig:velocity_over_time}
\end{figure}

%% file: img/viz_diversity.tex
\begin{figure}[t]
    \centering
    \subfloat[]{\fbox{\includegraphics[width=.23\textwidth, trim={5cm 6cm 5cm 4cm}, clip]{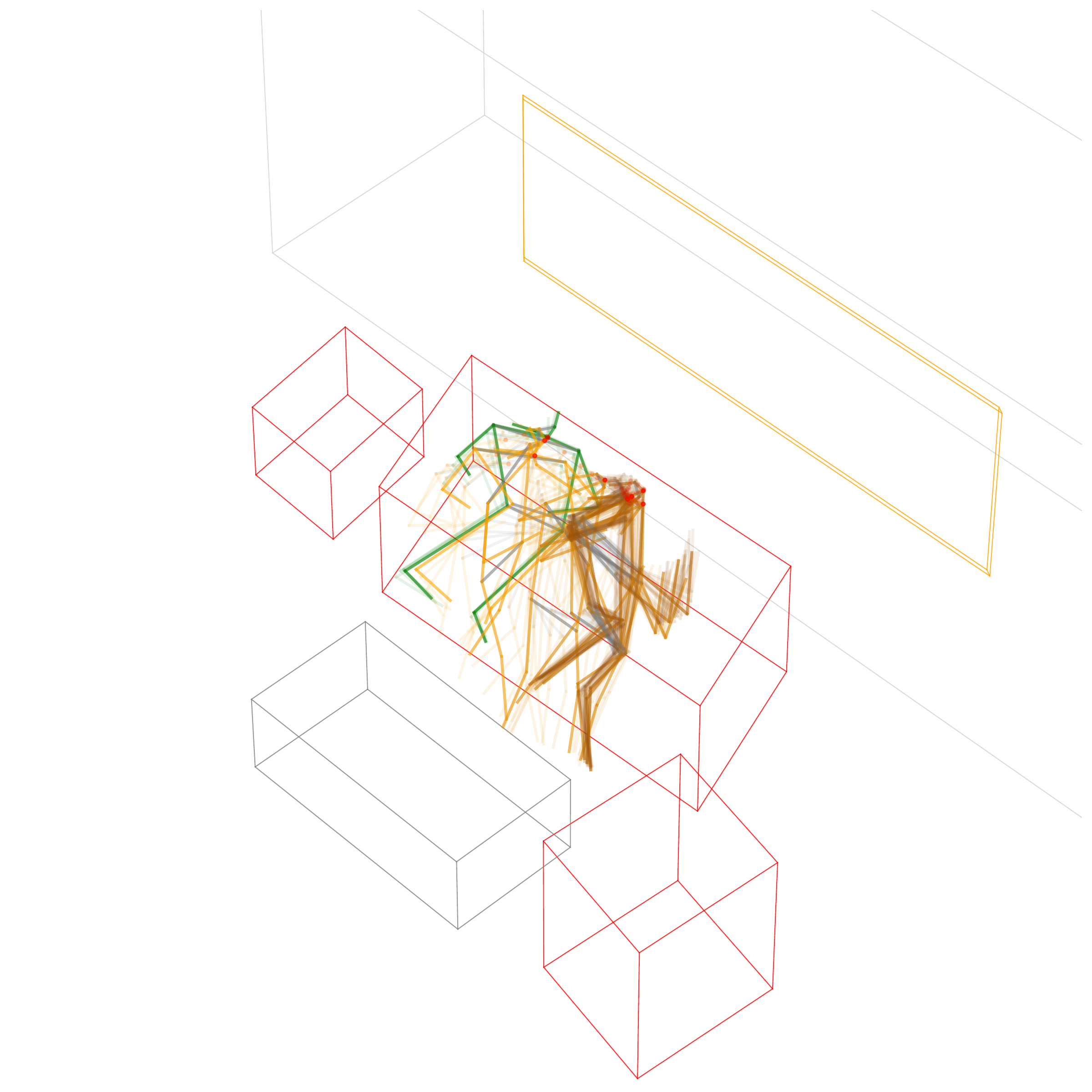}}}
    \subfloat[]{\fbox{\includegraphics[width=.23\textwidth, trim={5cm 6cm 5cm 4cm}, clip]{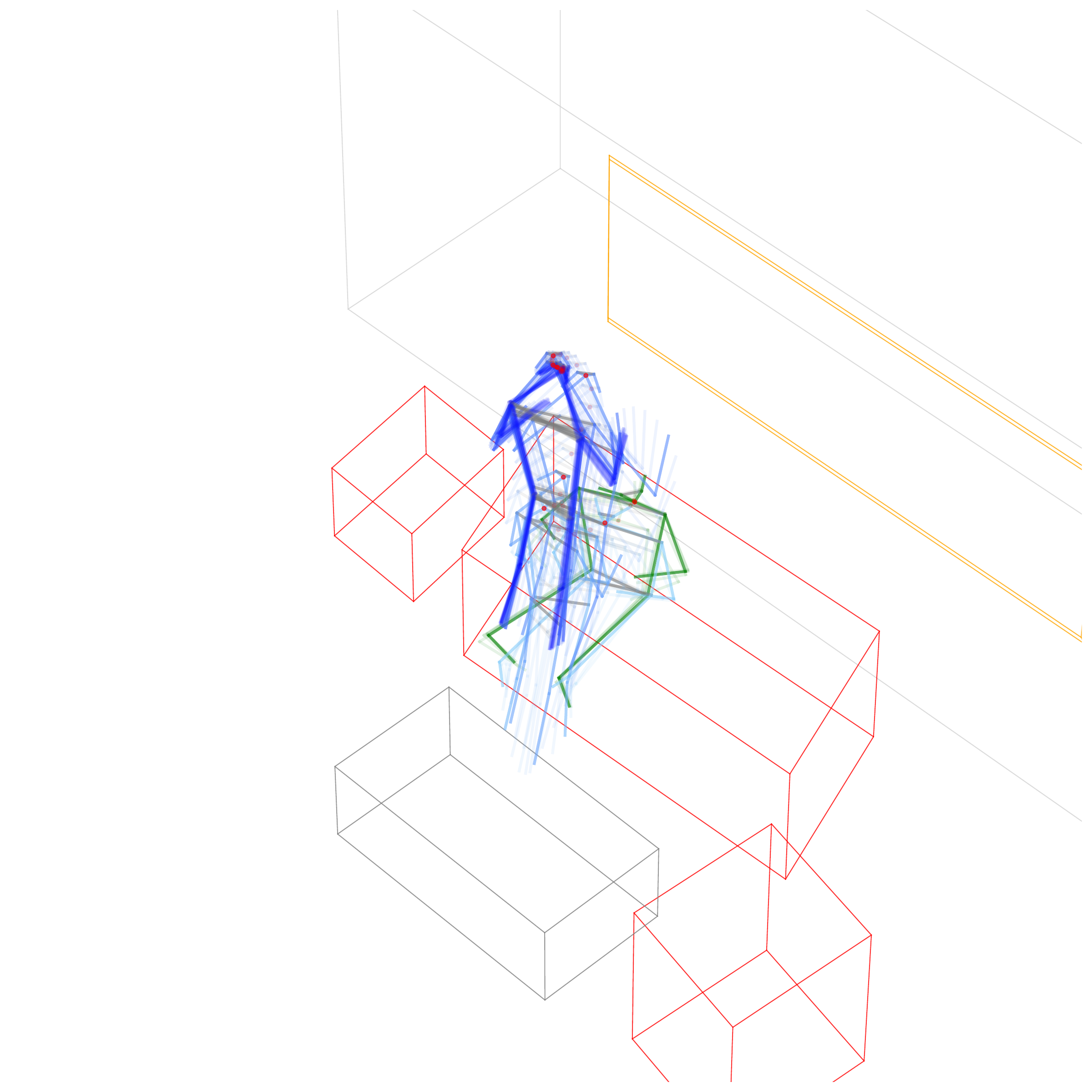}}}
\subfloat[]{\fbox{\includegraphics[width=.23\textwidth, trim={5cm 6cm 5cm 4cm}, clip]{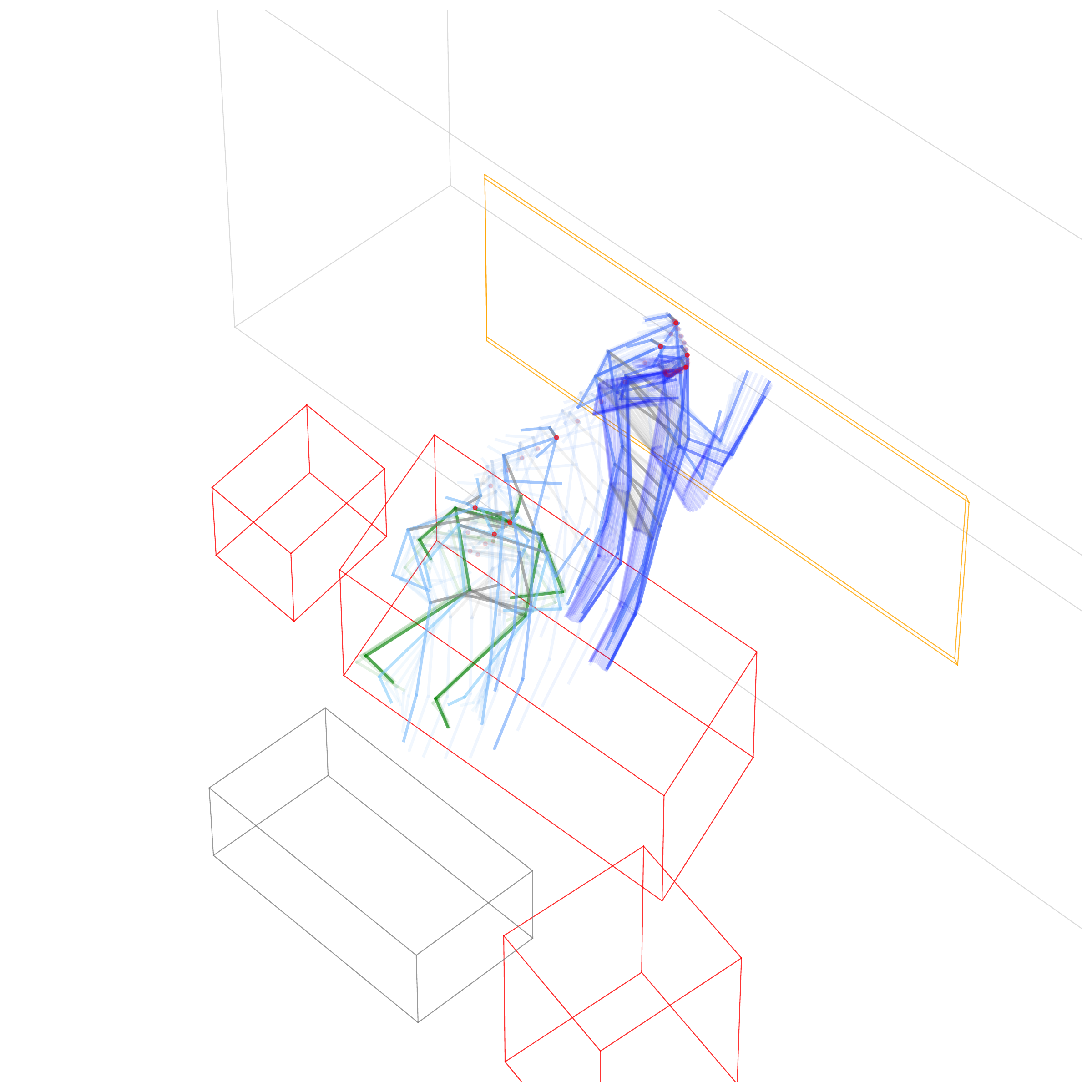}}}
\subfloat[]{\fbox{\includegraphics[width=.23\textwidth, trim={5cm 6cm 5cm 4cm}, clip]{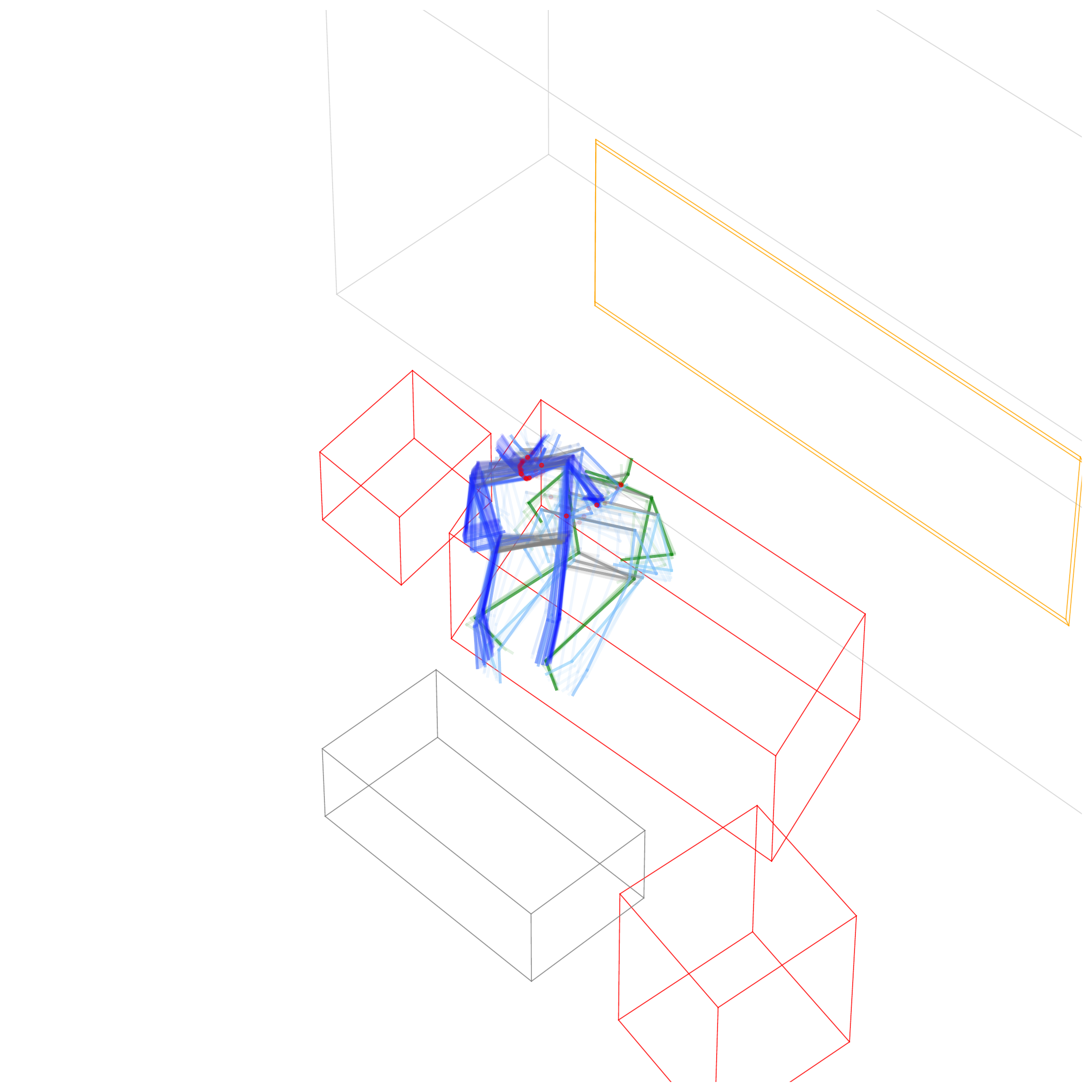}}}

    \caption{Samples of \ourUnder{} creating diverse motion based on a fixed input. In the input sequence \csquare{mplgreen}, a person starts to stand up. \textbf{a)} In the ground truth (\csquare{mplorange} to \csquare{mplsaddlebrown}), the person kneels on the sofa \csquare{mplred} to write on the whiteboard \csquare{mplorange}. \textbf{b--d)} \ourUnder{} (\csquare{mpllightskyblue} to \csquare{mplblue}) predicts writing on the whiteboard twice, once stepping on and once stepping over the sofa. The third prediction shows hesitant standing up motion.}
    \label{fig:viz_diversity}
\end{figure}

%% file: tables/ablations.tex
\begin{table}[]
    \centering
        \caption{Our model with scene context and other person encoder ablated. Best result or closest mean trajectory length to the ground truth is bold, the second best is underlined. Note that the evaluation metrics do not explicitly take human-human and human-scene interactions into account. We thus also provide a quick overview on perceived visual quality, more details in Section~\ref{sec:ablations}}
    \label{tab:ablations}
    \begin{tabular}{lrrrrrc}
    \toprule
        &&& \multicolumn{3}{c}{Trajectory Dist.} & Visual \\
        \cmidrule(lr){4-6}
        & NDMS $\uparrow$& UMWR $\uparrow$ & Mean & Std & $W_1$ $\downarrow$ & Quality \\
        \midrule
        Ground Truth &-&-& 1.17 & 1.70 & - & \cmark \cmark \\
        \midrule
        Ours &\textbf{0.17}&\textbf{0.20}& \underline{0.86} & 0.62 & \underline{0.57} & \cmark \\ 
        \midrule
w/o Scene Context &0.14&0.16& 0.63 & 0.51 & 0.59 & \mmark \\
w/o Other People &\underline{0.16}&\underline{0.19}& \textbf{0.93} & 0.98 & \textbf{0.36} & \cmark \\
\bottomrule
    \end{tabular}

\end{table}

%% file: sections/5_conclusion.tex
In this work, we propose our scene-aware social transformer model (SAST) that forecasts long-term human motion conditioned on both multi-person interactions and scene context. Our architecture combines causal convolutional pose encoder and decoder with a Transformer-based bottleneck that allows to model interactions between objects and persons on a far larger scale than previously done. We ablate these context information and see that both scene and multi-person context provide relevant information for realistic multi-person motion generation. Our inference procedure can produce highly-synchronized interdependent motion without the need for jointly forwarding all input sequences through the model, which is invaluable for large or varying numbers of people in a scene. We evaluate our approach on the Humans in Kitchens which requires modelling of widely varying numbers of people and objects in a scene and achieve very good performance compared to other approaches. However, our model still has limitations: The continuity between input and predicted sequence can be improved and limb movement realism drops during long global motion.

\paragraph{Future Work} 

To model realistic long-term human motion, we need to efficiently combine all available context information. We should work toward flexible multi-modal models that are able to reconcile more context information and guiding signals like objects, action labels, speech, etc.

Even though human-human and human-scene interactions are a central component of realism for human observers, current realism metrics do not capture those aspects. To purposefully work toward realistic long-term motion forecasting, we need to improve evaluation protocols to also include environment interactions.

\section*{Acknowledgments}

This publication was funded by the Deutsche Forschungsgemeinschaft (DFG, German Research Foundation) - Project-ID 454648639 - SFB 1528, 1927/5-2 (FOR 2535 Anticipating Human Behavior), and the ERC Consolidator Grant
FORHUE (101044724).

%% file: suppmat/videos.tex
\noindent \textbf{Please find example videos and information on the user study at \url{https://github.com/felixbmuller/SAST}} \\

\noindent The videos in our supplementary material are structured as follows:

\begin{description}
    \item[User Study By Model] All videos used in the user study, sorted by models (During the actual user study, they were sorted by categories, \eg coffee machine, sitting down, fridge, and not by model)
    \item[Examples Ours] Example outputs representative of our model. We show multiple outputs for the same input to highlight that our model allows to directly sample multiple realistic continuations (same file name, followed by \texttt{\_sX}
    \item[Ablations] Videos of our NoScene and NoOthers ablations, along with the output of our base model on the same input sequence.
    \item[Baselines] Example outputs representative of the baseline models 
\end{description}

Most filenames follow the schema 

\begin{quote}
\texttt{CATEGORY\_NUMBER[\_sX]\_[Model]\_Text.mp4}. 
\end{quote}

\noindent This means 
that the video was generated using the \texttt{NUMBER}$^\text{th}$ input sequence of the evaluation category \texttt{CATEGORY}. This uniquely identifies an input sequence, \ie samples with the same \texttt{CATEGORY\_NUMBER} were generated on the same input. \texttt{\_sX} is a running number, if we sampled multiple outputs for the same input.

\textbf{Note:} All videos show 1 second of input motion followed by 10 seconds of output motion. Videos with a runtime of 6 seconds are double speed.

\begin{table}[]
    \centering
    \begin{tabular}{lrr}
    \toprule
         & \multicolumn{2}{c}{Body Half} \\
         \cmidrule(lr){2-3}
         & Left & Right \\
         \midrule
         Ground Truth &  cornflowerblue \csquare{mplcornflowerblue} &  salmon \csquare{mplsalmon} \\
         Predicted & green \csquare{mplgreen} & orange \csquare{mplorange} \\
         \bottomrule
    \end{tabular}
    \caption{Color scheme for poses in videos}
    \label{tab:color_scheme}
\end{table}

\input{img/sceneD}

%% file: img/sceneD.tex
\begin{figure}
    \centering
    {\fbox{\includegraphics[width=\textwidth, trim={10cm 12cm 7cm 12.4cm},clip]{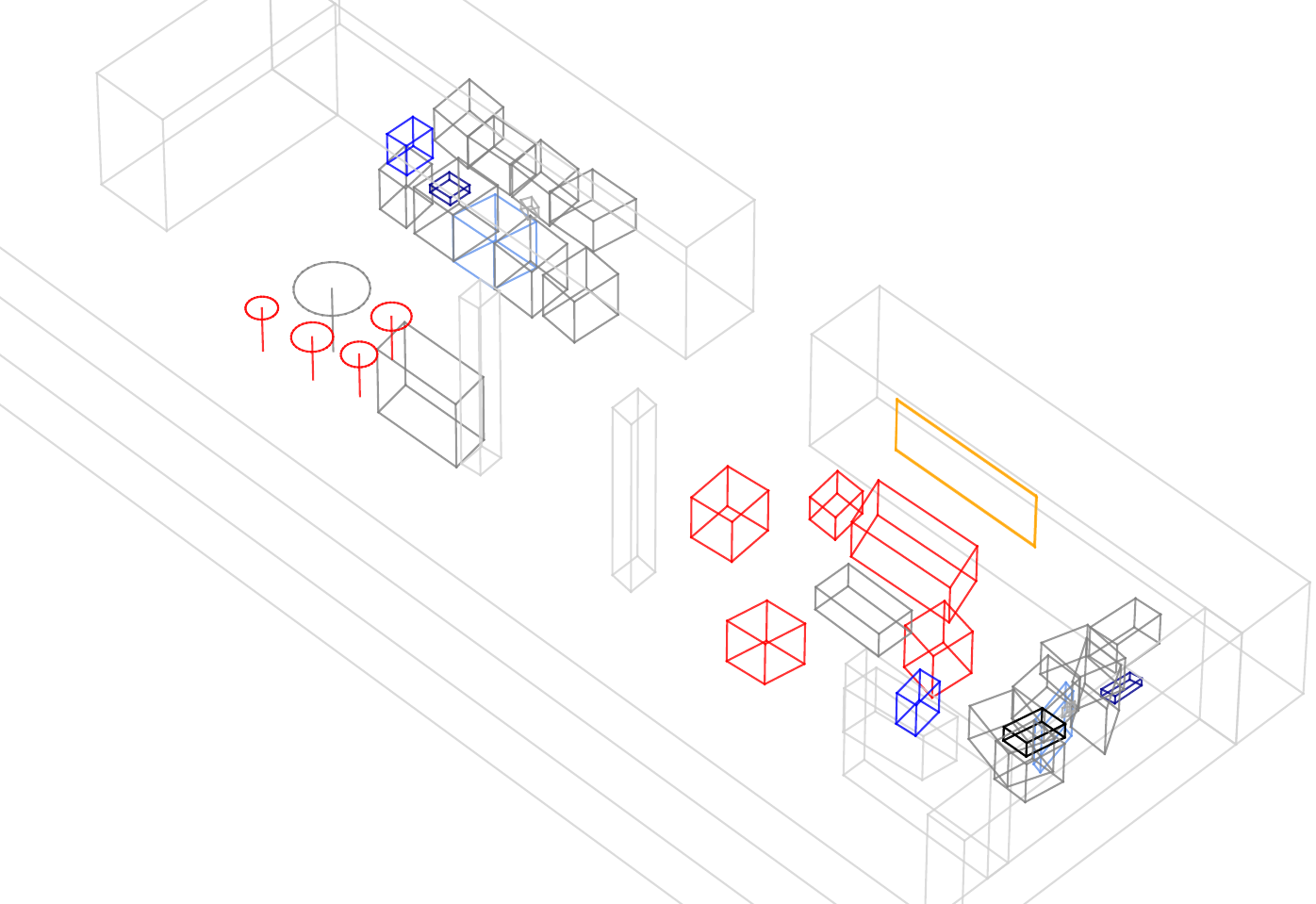}}}
    \caption{Scene geometry of the kitchen of the evaluation set D. Walls are light gray \csquare{mpllightgrey}. Tables, drawers, and cupboards are gray \csquare{mplgrey}. Chairs and sofas are red \csquare{mplred}. Less common objects (present at least once in each kitchen) include whiteboards (orange \csquare{mplorange}), coffee machines (blue \csquare{mplblue}), dishwashers (light blue \csquare{mplcornflowerblue}), sinks (dark blue \csquare{mpldarkblue}), and microwaves (black \csquare{black}). The figure does not cover the whole surface area per kitchen, but all objects are (partially) visible. There are 50 objects in this scene.}
    \label{fig:sceneD}
\end{figure}

%% file: suppmat/implementation_details.tex
\subsection{Our Architecture}

See Figure~\ref{fig:tcn_blocks_detail} for a detailed illustration of the encoder and decoder blocks. 
We choose a kernel size of 5 for the Pose Encoder and Pose Decoder, and a kernel size of 3 for the Others Encoder. We use 32 groups for group normalization throughout the model.

\input{img/tcn_blocks_detail}

\subsection{Realism Classifier}

To reduce input dimensionality, the first fully connected layer uses weight sharing, i.e. the same weights are applied for each body joint. Given a normalized single-person motion sequence $X = [j_{1,1}, j_{1, 2}, \dots, j_{J, 3}]$ with joint trajectories $j_{i,k} \in \Real^N$, \RealismClassifier{} calculates a realism score $s \in [0, 1]$ as
\begin{align*}
    j^*_{i,k} &= \DCT(j_{i,k}) & \forall j_{i,k} \in X\\
    e_i &= \operatorname{ReLU}(\operatorname{FC}^{(1)}(j^*_{i,1} \parallel j^*_{i,2} \parallel j^*_{i,3})) & \forall i \in \{1, \dots, J\} \\
    h &= \operatorname{ReLU}(\operatorname{FC}^{(2)}(e_{1} \parallel \dots \parallel e_J)) & \\
    s &= \operatorname{Sigmoid}(\operatorname{FC}^{(3)}(h)) &
\end{align*}

We choose a fixed sequence length of $N=50$ frames and a hidden dimensionality of 32 for $e_i$ and 512 for $h$. Thus, the \RealismClassifier{} has 284,000 parameters. We train the model using Adam with weight decay (source) and a learning rate of $10^{-3}$. We train for 6 epochs using a batch size of 16 and binary cross entropy loss. Our training data consists of 141,299 real sequences and 138,171 generated sequences of our model, SiMLPe, HisRep, TriPod, and MRT. 

During model evaluation, we use only model output that was not used during training of the \RealismClassifier{}.

%% file: img/tcn_blocks_detail.tex
\begin{figure}
    \centering
    \includegraphics[width=1.0\linewidth]{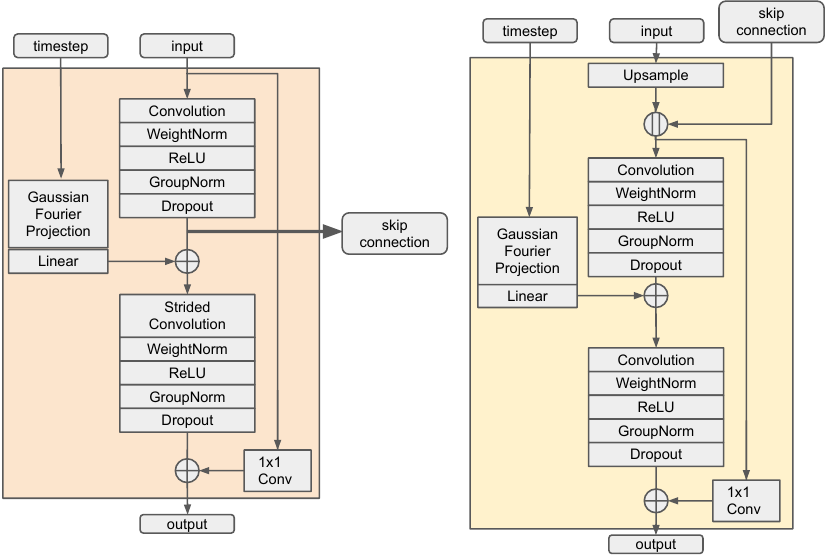}
    \caption{Detailed architecture of one encoder block \textbf{(left)} and decoder block \textbf{(right)}.}
    \label{fig:tcn_blocks_detail}
\end{figure}

%% file: suppmat/additional_metrics.tex
See Table~\ref{tab:realism_classifier} for detailed realism scores.

\input{tables/realism_classifier}

%% file: tables/realism_classifier.tex
\begin{table*}
    \centering
    \begin{tabular*}{\textwidth}{rllllllllll}
    \toprule
           &   \multicolumn{9}{c}{Realism Score $\times 100$ $\uparrow$} \\
         \cline{2-10}
          & 2s & 3s & 4s & 5s & 6s & 7s & 8s & 9s & 10s\\
          \midrule
         SiMLPe & \textbf{18.24} & \textbf{6.76} & \textbf{4.16} & \textbf{3.06} & \underline{2.40} & \underline{2.00} & \underline{1.71} & \underline{1.50} & \underline{1.33} \\
         MRT  & 4.23 & \underline{2.75} & 1.76 & 1.33 & 1.10 & 0.95 & 0.85 & 0.77 & 0.77 \\
         TriPod  & 3.39 & 0.64 & 0.37 & 0.27 & 0.22 & 0.19 & 0.17 & 0.15 & 0.14\\
         HisRep  & \underline{8.35} & 2.00 & 1.10 & 0.76 & 0.58 & 0.47 & 0.40 & 0.34 & 0.30 \\
         \midrule
         Ours  & 5.75 & 2.67 & \underline{2.80} & \underline{2.99} & \textbf{2.88} & \textbf{2.66} & \textbf{2.55} & \textbf{2.46} & \textbf{2.40} \\
         \bottomrule
    \end{tabular*}
    \caption{Realism scores for different models and output sequence lengths from two to ten seconds. For sequences longer than two seconds, we calculate the mean over all two-second subsequences with an offset of 5 frames.}
    \label{tab:realism_classifier}
\end{table*}